# Increasing the usefulness of already existing annotations through WSI registration

**Authors:**
Philippe Weitz[1], Viktoria Sartor[1], Balazs Acs[2,3], Stephanie Robertson[2], Daniel Budelmann[4], Johan Hartman[2,5], Mattias Rantalainen[1,5]

**Affiliations:**
1) Department of Medical Epidemiology and Biostatistics, Karolinska Institutet, Stockholm, Sweden
2) Department of Oncology and Pathology, Karolinska Institutet, Stockholm, Sweden
3) Department of Clinical Pathology and Cancer Diagnostics, Karolinska University Hospital, Stockholm, Sweden
4) Fraunhofer Institute for Digital Medicine MEVIS, Lübeck, Germany
5) MedTechLabs, BioClinicum, Karolinska University Hospital, Stockholm, Sweden

## Abstract

Computational pathology methods have the potential to improve access to precision medicine, as well as the reproducibility and accuracy of pathological diagnoses. Particularly the analysis of whole-slide-images (WSIs) of immunohistochemically (IHC) stained tissue sections could benefit from computational pathology methods. However, scoring biomarkers such as KI67 in IHC WSIs often necessitates the detection of areas of invasive cancer. Training cancer detection models often requires annotations, which is time-consuming and therefore costly. Currently, cancer regions are typically annotated in WSIs of haematoxylin and eosin (H&E) stained tissue sections. In this study, we investigate the possibility to register annotations that were made in H&E WSIs to their IHC counterparts. Two pathologists annotated regions of invasive cancer in WSIs of 272 breast cancer cases. For each case, a matched H&E and KI67 WSI are available, resulting in 544 WSIs with invasive cancer annotations. We find that cancer detection CNNs that were trained with annotations registered from the H&E to the KI67 WSIs only differ slightly in calibration but not in performance compared to cancer detection models trained on annotations made directly in the KI67 WSIs in a test set consisting of 54 cases. The mean slide-level AUROC is 0.974 [0.964, 0.982] for models trained with the KI67 annotations and 0.974 [0.965, 0.982] for models trained using registered annotations. This indicates that WSI registration has the potential to reduce the need for IHC-specific annotations. This could significantly increase the usefulness of already existing annotations.

## Introduction

Computational pathology methods have the potential to improve access to precision medicine, as well as the reproducibility and accuracy of pathological diagnoses. Some of these methods aim to automate current routine clinical workflows, such as Gleason grading of prostate cancer biopsies [1–3]. Other methods aim to predict diagnostically and prognostically relevant information that pathologists can typically not obtain from WSIs alone. Examples of this are the prediction of molecular subtypes [4, 5] or gene expression [6–9]. A commonality of these methods is that most of them detect invasive cancer regions as a first step. These regions are then used to train models for the corresponding objective and predictions are limited to these regions. Currently, most methods only analyse WSIs of H&E stained tissue sections. However, IHC staining for biomarker scoring is an essential component of clinical diagnostics. Biomarker scoring is typically performed within regions of invasive cancer. In order to detect these regions, stain-specific or stain-independent cancer detection models may be required. A limiting factor for training semantic segmentation models for computational pathology modelling pipelines is the generation of annotations by pathologists, which is time-consuming and therefore costly.

There are several studies that aim to mitigate this by using WSI registration. This allows the use of pathologist annotations or predicted semantic segmentations that are available in H&E WSIs to classify regions in IHC WSIs [10–12]. WSI registration is an active field of research. Recently, the ACROBAT challenge ([acrobat.grand-challenge.org/](acrobat.grand-challenge.org/)) was held in conjunction with MICCAI 2022. Its objective was to build on the findings from the ANHIR challenge [13] and expand its findings to data sets consisting of tissue slides that originate from routine diagnostic workflows. While it is a promising approach to register WSIs to transfer H&E segmentations for predictions from IHC WSIs, particularly if these segmentations can be predicted by a computer vision model, it relies on the availability of matched H&E and IHC WSIs at prediction time. These methods are therefore not applicable if no H&E stained tissue section is available that can be registered to the corresponding IHC. This might e.g. be the case if sections are not consecutive, since the registration performance of current methods is likely to deteriorate with increasing distance between the sections. Furthermore, it may be desirable to disentangle registration performance from the IHC model performance to increase robustness for clinical applications. This could be achieved by training stain-specific cancer detection models. However, there is a large number of IHC stains that can be of interest, which would require substantial annotation resources. A more promising approach might therefore be to register annotations that already exist for H&E WSIs to the IHC domain. This would allow to train stain-specific cancer detection models that are also applicable even if a sufficiently precise registration is not possible during prediction.

Currently, there is no investigation of the registration of annotations between stains to train models in the IHC domain. In this study, we therefore systematically investigate the registration of annotations between H&E and IHC WSIs for subsequent model training. This has the potential to increase the usefulness of already existing annotations without any additional cost.

## Materials & Methods

### Data set & splits

The data set used in this study consists of 544 WSIs from 272 female patients with primary breast cancer, and with two WSIs for each patient. The first WSI for each patient depicts an H&E-stained tissue section, while the second WSI contains a corresponding tissue section IHC stained for KI67. Paired sections were cut from the same formalin-fixed paraffin-embedded (FFPE) tumour block, but are not necessarily consecutive. All sections originate from routine clinical workflows from the time of first diagnosis. The study in whose context these WSIs were generated has received approval by the regional ethics review board (Stockholm - Ref. 2017/2106-31, Amendment: 2018/1462-32). All H&E WSIs were annotated by a pathologist specialising in breast pathology. H&E annotations include the classes invasive cancer (IC), ductal carcinoma in situ (DCIS), lobular carcinoma in situ (LCIS), non-malignant changes, artefacts, lymphovascular invasion and tissue. Annotations for the KI67 WSIs were generated by a different pathologist who specialises in breast pathology. For these WSIs, annotations of invasive cancer areas are available, which likely also include some areas of DCIS. The H&E WSIs were solely used for WSI registration to transfer H&E annotations to the KI67 WSIs and for tissue detection in order to exclude control tissue in the KI67 WSIs. No H&E tissue was used for model training.

The 272 cases were split into a development set that consists of 218 cases (80.1%) and a test set that consists of 54 cases (19.9%). The development set was further split into 5 cross-validation (CV) folds for hyperparameter tuning. The training data of each CV fold was further split into a tune set (15%) for early stopping and data that the model was fit to (85%). Data splits at all levels were conducted on the patient-level and stratified for clinical KI67 scores that were obtained from medical records. KI67 scores are a number in [0, 100] that indicates the percentage of KI67-positive cancer cells within selected hot spots. For the stratification, cases were assigned to a KI67-low and a KI67-high group, split on the median KI67 score (25%) in the development data set. For 21 out of the 272 cases, the clinical KI67 score was not available. These 21 cases were randomly distributed to the KI67-low or KI67-high group for stratification. The distribution of KI67 scores among the different data splits is depicted in Supplementary Figure 1.

## WSI registration

WSI image pairs were registered with the WSI registration algorithm that the team AGHSSO used in the ACROBAT WSI registration challenge. It is based on the DeepHistReg WSI registration method [14], which was initially developed for the ANHIR WSI registration challenge [13] and then adjusted for the more realistic data set from the ACROBAT challenge. The registration of an image pair yields a deformation field that can be used to transfer coordinates from the source to the target image. All annotations in this study were generated in QuPath [15] and consist of polygons that indicate a specific class. The deformation fields found during the WSI registration were used to transform the coordinates of the vertices of these polygons to the coordinate systems of the matched KI67 WSIs. We then generated invasive cancer masks with a resolution of 7.264 µm/pixel based on these polygon coordinates. An example of a H&E WSI before and after registration, as well as the target KI67 WSI, is depicted in Figure 1.

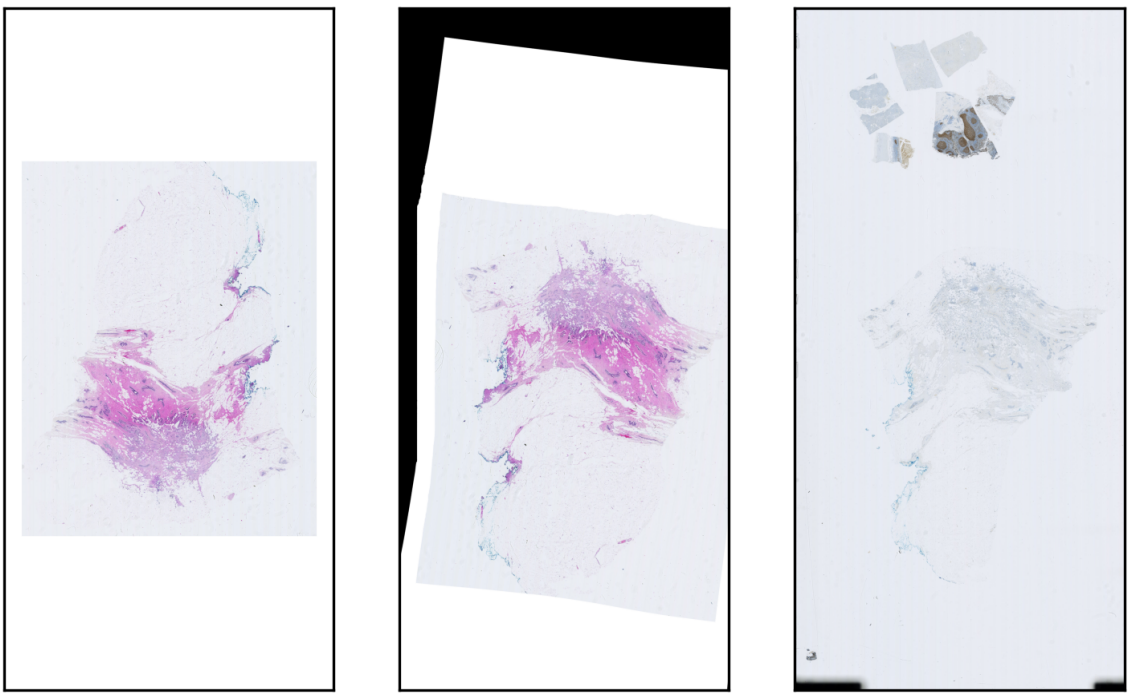

**Figure 1.** *Randomly selected example of a H&E WSI before and after registration, as well as the target KI67 WSI. The KI67 WSI contains control tissue patches, which we aimed to exclude during image preprocessing.*

## Image preprocessing

The image preprocessing in this study consists of two steps, tissue detection and generation of high-resolution image patches. For tissue detection, we re-implemented the method described in [16] in PyTorch. Bándi et al. published ten WSIs and corresponding tissue annotations with tissue of different organs and stains. Additionally, we annotated tissue in 15 randomly selected WSIs from the publicly available ACROBAT data set [17], three from each available stain (H&E, ER, PGR, HER2, KI67). We performed the tissue detection at a resolution of 3.64 µm/pixel, with all other parameters as described in [16], and applied it to all 544 WSIs in this study. The resulting tissue masks were then post-processed by removing salt-and-pepper noise and by removing contiguous tissue regions that have more than 50% of their area in the 10% outer edge of the WSI, as these were mainly artefacts. We then counted the number of contiguous tissue regions in the H&E WSIs and selected the same number of tissue regions in the matched KI67 WSIs to exclude areas with control tissue. An example of a KI67 WSI with control tissue is shown in Figure 1.

WSIs were then tiled at 20X magnification (0.454 µm/pixel) into quadratic patches with an edge length of 598 pixels with a stride of 598 pixels in each direction. For a patch to be generated, it needed to contain at least 50% tissue pixels based on the tissue masks. This yielded 1,136,400 image patches in total.

Each image patch was then assigned a binary label for invasive cancer based on the annotations performed in the KI67 WSIs. A second invasive cancer label based on the registered H&E annotations was assigned to image patches. For all annotations and patches, an image patch was assigned a positive label if at least half of all pixels in a patch belong to the invasive cancer class. Registered annotations yield 298,654 invasive cancer tiles, whereas there are 345,494 cancer tiles that originate directly from the KI67-annotator. The main reason for this difference in number of cancer tiles is likely the partial presence of DCIS in the IC regions in the KI67 annotations, but can also be a consequence of inter-observer variability, which may be enhanced through the different stains.

## Model training and prediction

Hyperparameters were optimised using CV with the previously described data splits in the development data. Final models were then fit to the entire development data based on these parameters. We did not explore a large hyperparameter space but focused the grid search on ranges of values that we know to work well for training IHC cancer detection CNNs from previous work. We compared the ResNet18 and InceptionV3 model architectures initialised with ImageNet weights. Models were trained with mixed precision using PyTorch 1.13.0 and RayTune 2.1.0 on a computer with Ubuntu 18.04.6 and four RTX 2080Ti GPUs. We tuned the learning rate in [0.001, 0.0005] and modified the learning rate schedule based on the tuning loss curves. Models were trained in partial epochs consisting of 10,000 images during training and 5,000 images during performance evaluation at the end of each partial epoch. The batch size was set to 32. Batches both for training and validation were generated by balanced sampling between the two classes. Early stopping was performed based on the tuning data loss. The learning rate was multiplied with 0.2 if there was no improvement in the tuning loss for 12 partial epochs, while the training was stopped if there was no improvement for 25 partial epochs. Training images were randomly augmented with Albumentations 1.3.0 by applying random mirroring and 90° rotations, adjusting the brightness in the range [0.9, 1.1], the contrast in the range [0.9, 1.1], the hue in the range [-0.1, 0.1], the saturation in the range [0.9, 1.1], additive Gaussian noise with a variance of 12.75 and blurring images with a Gaussian kernel size in [3, 7]. We then predicted the respective validation folds, concatenated predictions across validation folds for all models and computed performance metrics with the respective labels. The results are available in Supplementary Table 1. While all trained models seem to perform very similarly, models with higher learning rate performed slightly better. The difference between ResNet18 and InceptionV3 appeared negligible, we therefore chose ResNet18 in order to reduce the computational cost. Investigating the tune loss curves also informed the number of training epochs and learning rate schedule for the final model training. Models that were optimised against the registered labels were trained for 100 partial epochs, while reducing the learning rate with a factor of 0.2 at 40 and 80 partial epochs. Models that were optimised with the IHC-specific KI67 labels were trained for 120 partial epochs, with a corresponding reduction in the learning rate at 50 and 100 partial epochs. Classification thresholds for each combination of architecture and hyperparameters were established using the validation data based on the Youden Index and are available in Supplementary Table 1. These thresholds were later used to binarise the test data predictions. The ResNet18 models that we ultimately evaluated in the test set were then trained with a learning rate of 0.001 using data from the entire development data set, without further monitoring a tuning loss. We trained ten models against each the registered and the IHC-specific labels.

Prediction of test data yields a number in [0, 1] for each of the ten base models in the respective ensemble. We averaged these predictions for each tile and computed an AUROC for each WSI in the test set. We then applied the previously established classification thresholds and generated invasive cancer masks at a resolution of 7.264 μm/pixel and removed salt-and-pepper noise. The resulting masks were then used to compute performance metrics that require binarised predictions.

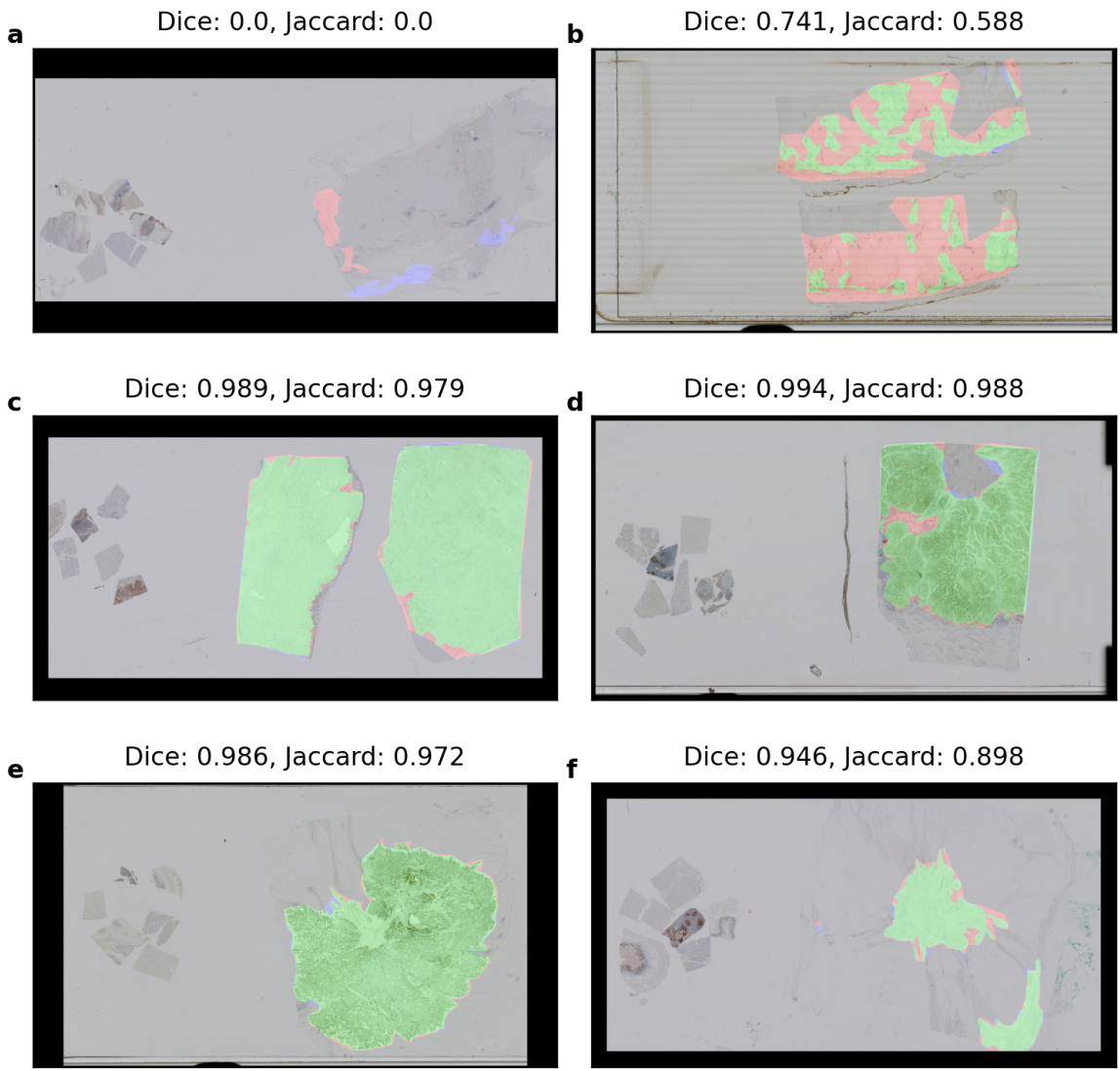

**Figure 2.** *Comparison of IHC-specific and registered annotations (left column) and predictions from models trained with IHC-specific or registered annotations (right column). Green indicates areas where both indicate invasive cancer, red where only the IHC-specific annotations or the respectively trained model indicate invasive cancer and blue where only the registered annotations or the respectively trained model indicate invasive cancer. a) shows the WSI with the lowest agreement (Dice) between annotations, b) the WSI with the lowest agreement between predictions, c) the WSI with the highest agreement between annotations, d) the WSI with the highest agreement between predictions, e) a randomly selected WSI with a pair of annotations and f) a randomly selected WSI with a pair of predictions.*

# Results

## Analysis of inter-annotator and inter-model agreement

First, we analysed the overlap of IHC-specific annotations that were directly generated in the KI67 WSIs and the annotations that were generated in H&E WSIs and then registered to the KI67 WSIs. Figure 2 shows examples of lowest and highest agreement between annotators and respectively trained models, as well as randomly selected WSIs. Annotators were compared in the entire data set of 272 WSIs, whereas predictions were only compared in the test set. The mean Dice coefficient between annotations is 0.826 [0.795, 0.856] and the mean Jaccard index is 0.76 [0.728, 0.791]. There are 15 WSIs in the data set for which the Dice coefficient between IHC-specific and registered annotations is 0. The predictions of the two modelling approaches are more similar, with a mean Dice coefficient of 0.916 [0.898, 0.933] and a mean Jaccard index of 0.852 [0.824, 0.879]. The lowest agreement between models results in a Dice coefficient of 0.741 and a Jaccard index of 0.588, as shown in Figure 2 b).

## Analysis of model performance

Performance metrics were computed on the WSI-level with the annotations that were generated directly in the KI67 WSIs as the ground truth for both training approaches. This results in 54 values for each metric, one for each case in the test set. Table 1 lists the means across these 54 values, as well as 95% confidence intervals, which were computed using 10,000 bootstrap samples. The AUROC as the metric that is least dependent on calibration appears to be the most similar, with a mean value of 0.974 for both ensembles. We also tested the distributions of the metrics for differences using paired Wilcoxon signed-rank tests [18]. The resulting p-values were adjusted for multiple comparisons with the method described by Benjamini and Hochberg (BH, [19]) and listed in the last row of Table 1. While the distributions of AUROCs, Dice coefficients, Jaccard similarities and F1 scores are not distinguishable, there are differences in the distributions of the more calibration-dependent specificities, sensitivities/recalls and precisions. However, while these differences are statistically significant, the scales of the differences are small. Figure 3 a) - h) shows scatterplots of the metric values for each of the 54 test set cases for slide-wise metric values between the two modelling approaches. Figure 3 i) - l) depicts scatterplots for selected metrics for both ensembles against the KI67-score for the 49 test set cases where the score is available.

|  | AUROC | Dice | Jaccard | Accuracy | F1 | Specificity | Sensitivity/ Recall | Precision |
|---|---|---|---|---|---|---|---|---|
| Training with IHC-specific annotations | 0.974 [0.964, 0.982] | 0.816 [0.768, 0.858] | 0.718 [0.662, 0.771] | 0.919 [0.899, 0.936] | 0.816 [0.767, 0.858] | 0.921 [0.898, 0.94] | 0.915 [0.882, 0.94] | 0.78 [0.72, 0.835] |
| Training with registered annotations | 0.974 [0.965, 0.982] | 0.813 [0.765, 0.858] | 0.716 [0.657, 0.769] | 0.921 [0.9, 0.939] | 0.813 [0.764, 0.857] | 0.931 [0.908, 0.951] | 0.888 [0.851, 0.917] | 0.798 [0.737, 0.853] |
| BH-adj. p-value | 0.962 | 0.879 | 0.885 | 0.891 | 0.879 | 0.006 | <0.001 | 0.017 |

**Table 1.** *Mean and 95% confidence intervals for slide-wise performance metrics for models that were trained with IHC-specific annotations that were generated using KI67 WSIs, as well as for models that were trained with annotations that were registered from H&E to the KI67 WSIs. The IHC-generated KI67 annotations were used as the ground truth for all models for performance evaluation. The last row contains BH-adjusted p-values from paired Wilcoxon signed-rank tests between the distributions of performance metrics from the first two rows.*

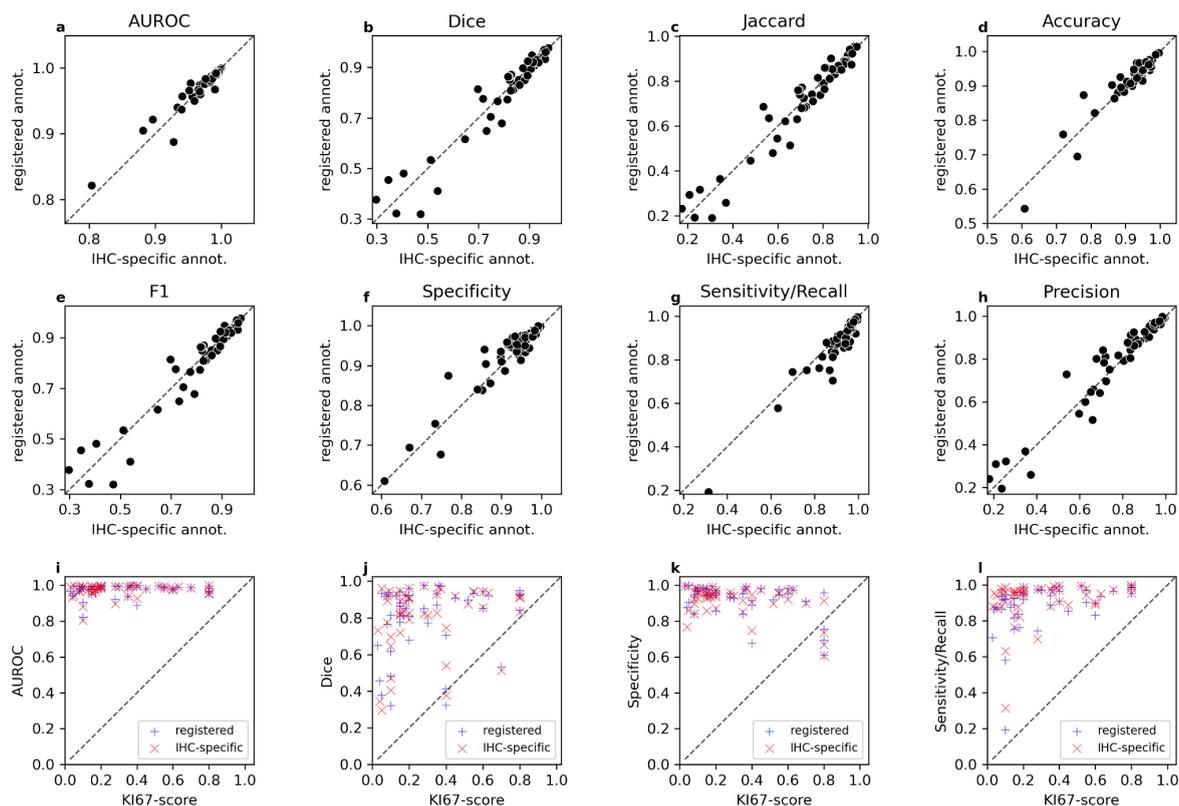

**Figure 3.** *Scatterplots of slide-wise performance metrics for models trained with IHC-specific annotations in the KI67 WSIs compared to values for models trained with registered annotations and KI67-scores. Dashed lines indicate the diagonal between 0 and 1. a) - f) show a comparison of the performance metrics between the models for the 54 test set cases. i) - l) depict selected performance metrics compared to clinical KI67-scores for 49 test set cases where scores are available.*

We also investigated the Spearman correlation between metrics for each modelling approach and clinical KI67-scores in this subset. All correlations are available in Supplementary Table 2. The only metric for which the 95% confidence interval of the Spearman correlation does not include 0 for both modelling approaches is the sensitivity/recall, with a Spearman correlation of 0.389 [0.135, 0.611] and 0.477 [0.227, 0.682] for training with IHC-specific and registered labels respectively. A scatterplot for sensitivity/recall compared to the KI67-score is available in Figure 3 l). This indicates an increase in sensitivity/recall for WSIs with a higher KI67-score. Notably, there is also a weaker negative correlation between the KI67-score and specificity, with -0.225 [-0.533, 0.114] and -0.256 [-0.547, 0.07] for IHC-specific and registered labels respectively, shown in Figure 3 k).

## Discussion

In this study, we investigated the possibility to train cancer detection models based on annotations that were generated by a pathologist in H&E WSIs and then transferred to IHC WSIs using a WSI registration algorithm. We compared these annotations and resulting model performances to a cancer detection model that was trained on annotations generated directly for the KI67 WSIs.

When comparing registered annotations to those that were generated directly with the KI67 WSIs, the Dice coefficient appears to be relatively low, indicating a high inter-assessor variability. This may not be surprising considering that WSIs of differently stained tissue were used for the annotations, and that the paired tissue sections were not necessarily consecutive. In the case of distant sections, IC regions may appear in different regions of the WSI. Potential registration errors could further increase these differences, again particularly for distant sections. However, this appears to only moderately impact model training, which is shown by the higher Dice coefficients between the predictions of the two modelling approaches. This indicates that the label noise during model training only has a minor impact on the model predictions. This is also supported by the high similarity of performance metrics between the two modelling approaches. Particularly the calibration-independant AUROC appears to be practically identical. This strongly supports the hypothesis that it is possible to train cancer detection models based on registered annotations without loss in performance. However, even metrics that do require binarised predictions are largely not distinguishable between the modelling approaches, with the exceptions of specificity, sensitivity/recall and precision, which depend on calibration the most. While testing for differences in distributions of these metrics indicates that they are different, the sizes of these differences are small. Nevertheless, it could be beneficial to annotate a small calibration set of WSIs in the target IHC stain after training a detection model based on registered annotations.

When correlating slide-wise performance metrics with clinical KI67-scores, the sensitivity/recall has the highest correlation. This may not be surprising, as the positively stained cancer cells are visually easily distinguishable from surrounding benign tissue. There is also a weaker negative correlation between specificity and KI67-scores. This could be explained by a failure of the detection models to distinguish IC from DCIS, which can also contain cells with a high KI67-expression.

This also points to some limitations of this study. According to the pathologist who annotated the KI67 WSIs, the annotations are not completely free of DCIS. The impact of this label noise during model training is likely smaller than the impact during performance evaluation, but this is difficult to quantify. Furthermore, we only investigated a magnification of 20X, which is well suited for invasive cancer detection, but not optimal for distinguishing IC and DCIS due to the lack of context [20]. Multi-resolution models might be required to further improve model performance. It is also possible that with improved performance, differences between training with registered and IHC-specific annotations become apparent. During future studies, we would therefore like to include additional IHC stains, increase the number of investigated classes to include e.g. DCIS and perform multi-resolution modelling.

The results that we obtained in this study however show that it is possible to train highly accurate invasive cancer detection models with registered annotations and that these models are not inferior to models trained with IHC-specific KI67 annotations. We therefore conclude that WSI registration has the potential to significantly increase the usefulness of already existing H&E annotations at no additional cost.

# Code availability

Code for the WSI-registration used in this study is available from https://github.com/MWod/DeeperHistReg-ACROBAT-submission. A Tensorflow implementation of the IHC tissue detection is available from https://peerj.com/articles/8242/. Code for model training depends on internal tooling and can therefore not be shared easily.

# Author contributions

P.W. and M.R. conceptualised the study. P.W. drafted this manuscript. P.W. curated and preprocessed the data set. D.B. contributed code for control tissue removal. S.R. and B.A. annotated the data set. P.W. and V.S. registered WSIs. P.W. optimised models and performed the statistical analysis. M.R. and J.H. acquired funding and organised data collection. All authors contributed to editing this manuscript.

## Acknowledgements

We acknowledge Nguyen Thuy Duong Tran for support with digitising histopathology slides. We acknowledge funding from: Vetenskapsrådet (Swedish Research Council), Cancerfonden (Swedish Cancer Society), ERA PerMed (ERAPERMED2019-224-ABCAP), MedTechLabs, Swedish e-science Research Centre (SeRC), VINNOVA and SweLife.

## References

1. Ström, P., Kartasalo, K., Olsson, H., Solorzano, L., Delahunt, B., Berney, D.M., Bostwick, D.G., Evans, A.J., Grignon, D.J., Humphrey, P.A., Iczkowski, K.A., Kench, J.G., Kristiansen, G., van der Kwast, T.H., Leite, K.R.M., McKenney, J.K., Oxley, J., Pan, C.-C., Samaratunga, H., Srigley, J.R., Takahashi, H., Tsuzuki, T., Varma, M., Zhou, M., Lindberg, J., Lindskog, C., Ruusuvuori, P., Wählby, C., Grönberg, H., Rantalainen, M., Egevad, L., Eklund, M.: Artificial intelligence for diagnosis and grading of prostate cancer in biopsies: a population-based, diagnostic study. Lancet Oncol. 21, 222–232 (2020).
2. Bulten, W., Pinckaers, H., van Boven, H., Vink, R., de Bel, T., van Ginneken, B., van der Laak, J., Hulsbergen-van de Kaa, C., Litjens, G.: Automated deep-learning system for Gleason grading of prostate cancer using biopsies: a diagnostic study. Lancet Oncol. 21, 233–241 (2020).
3. Bulten, W., Kartasalo, K., Chen, P.-H.C., Ström, P., Pinckaers, H., Nagpal, K., Cai, Y., Steiner, D.F., van Boven, H., Vink, R., Hulsbergen-van de Kaa, C., van der Laak, J., Amin, M.B., Evans, A.J., van der Kwast, T., Allan, R., Humphrey, P.A., Grönberg, H., Samaratunga, H., Delahunt, B., Tsuzuki, T., Häkkinen, T., Egevad, L., Demkin, M., Dane, S., Tan, F., Valkonen, M., Corrado, G.S., Peng, L., Mermel, C.H., Ruusuvuori, P., Litjens, G., Eklund, M., PANDA challenge consortium: Artificial intelligence for diagnosis and Gleason grading of prostate cancer: the PANDA challenge. Nat. Med. 28, 154–163 (2022).
4. Kather, J.N., Heij, L.R., Grabsch, H.I., Loeffler, C., Echle, A., Muti, H.S., Krause, J., Niehues, J.M., Sommer, K.A.J., Bankhead, P., Kooreman, L.F.S., Schulte, J.J., Cipriani, N.A., Buelow, R.D., Boor, P., Ortiz-Brüchle, N., Hanby, A.M., Speirs, V., Kochanny, S., Patnaik, A., Srisuwananukorn, A., Brenner, H., Hoffmeister, M., van den Brandt, P.A., Jäger, D., Trautwein, C., Pearson, A.T., Luedde, T.: Pan-cancer image-based detection of clinically actionable genetic alterations. Nature Cancer. 1, 789–799 (2020).
5. Schaumberg, A.J., Rubin, M.A., Fuchs, T.J.: H&E-stained Whole Slide Image Deep Learning Predicts SPOP Mutation State in Prostate Cancer, https://www.biorxiv.org/content/10.1101/064279v9, (2018). https://doi.org/10.1101/064279.
6. Wang, Y., Kartasalo, K., Weitz, P., Ács, B., Valkonen, M., Larsson, C., Ruusuvuori, P., Hartman, J., Rantalainen, M.: Predicting Molecular Phenotypes from Histopathology Images: A Transcriptome-Wide Expression-Morphology Analysis in Breast Cancer. Cancer Res. 81, 5115–5126 (2021).
7. Schmauch, B., Romagnoni, A., Pronier, E., Saillard, C., Maillé, P., Calderaro, J., Kamoun, A., Sefta, M., Toldo, S., Zaslavskiy, M., Clozel, T., Moarii, M., Courtiol, P., Wainrib, G.: A deep learning model to predict RNA-Seq expression of tumours from whole slide images. Nat. Commun. 11, 3877 (2020).
8. Fu, Y., Jung, A.W., Torne, R.V., Gonzalez, S., Vöhringer, H., Shmatko, A., Yates, L.R., Jimenez-Linan, M., Moore, L., Gerstung, M.: Pan-cancer computational histopathology reveals mutations, tumor composition and prognosis. Nature Cancer. 1, 800–810 (2020).
9. Weitz, P., Wang, Y., Kartasalo, K., Egevad, L., Lindberg, J., Grönberg, H., Eklund, M., Rantalainen, M.: Transcriptome-wide prediction of prostate cancer gene expression from histopathology images using co-expression-based convolutional neural networks. Bioinformatics. 38, 3462–3469 (2022).
10. Huang, Z., Shao, W., Han, Z., Alkashash, A.M., De la Sancha, C., Parwani, A.V., Nitta, H., Hou, Y., Wang, T., Salama, P., Rizkalla, M., Zhang, J., Huang, K., Li, Z.: Artificial intelligence reveals features associated with breast cancer neoadjuvant chemotherapy responses from multi-stain histopathologic images. NPJ Precis Oncol. 7, 14 (2023).
11. Swiderska-Chadaj, Z., Gallego, J., Gonzalez-Lopez, L., Bueno, G.: Detection of Ki67 Hot-Spots of Invasive Breast Cancer Based on Convolutional Neural Networks Applied to Mutual Information of H&E and Ki67 Whole Slide Images. NATO Adv. Sci. Inst. Ser. E Appl. Sci. 10, 7761 (2020).
12. Duanmu, H., Bhattarai, S., Li, H., Shi, Z., Wang, F., Teodoro, G., Gogineni, K., Subhedar, P., Kiraz, U., Janssen, E.A.M., Aneja, R., Kong, J.: A spatial attention guided deep learning system for prediction of pathological complete response using breast cancer histopathology images. Bioinformatics. 38, 4605–4612 (2022).
13. Borovec, J., Kybic, J., Arganda-Carreras, I., Sorokin, D.V., Bueno, G., Khvostikov, A.V., Bakas, S., Chang, E.I.-C., Heldmann, S., Kartasalo, K., Latonen, L., Lotz, J., Noga, M., Pati, S., Punithakumar, K., Ruusuvuori, P., Skalski, A., Tahmasebi, N., Valkonen, M., Venet, L., Wang, Y., Weiss, N., Wodzinski, M., Xiang, Y., Xu, Y., Yan, Y., Yushkevich, P., Zhao, S., Munoz-Barrutia, A.: ANHIR: Automatic Non-Rigid Histological Image Registration Challenge. IEEE Trans. Med. Imaging. 39, 3042–3052 (2020).
14. Wodzinski, M., Müller, H.: DeepHistReg: Unsupervised Deep Learning Registration Framework for Differently Stained Histology Samples. Comput. Methods Programs Biomed. 198, 105799 (2021).
15. Bankhead, P., Loughrey, M.B., Fernández, J.A., Dombrowski, Y., McArt, D.G., Dunne, P.D., McQuaid, S., Gray, R.T., Murray, L.J., Coleman, H.G., James, J.A., Salto-Tellez, M., Hamilton, P.W.: QuPath: Open source software for digital pathology image analysis. Sci. Rep. 7, 16878 (2017).
16. Bándi, P., Balkenhol, M., van Ginneken, B., van der Laak, J., Litjens, G.: Resolution-agnostic tissue segmentation in whole-slide histopathology images with convolutional neural networks. PeerJ. 7, e8242 (2019).
17. Weitz, P., Valkonen, M., Solorzano, L., Carr, C., Kartasalo, K., Boissin, C., Koivukoski, S., Kuusela, A., Rasic, D., Feng, Y., Pouplier, S.K.S., Sharma, A., Eriksson, K.L., Latonen, L., Laenkholm, A.-V., Hartman, J., Ruusuvuori, P., Rantalainen, M.: ACROBAT -- a multi-stain breast cancer histological whole-slide-image data set from routine diagnostics for computational pathology, http://arxiv.org/abs/2211.13621, (2022).
18. Wilcoxon, F.: Individual Comparisons by Ranking Methods. Biometrics Bulletin. 1, 80–83 (1945).
19. Benjamini, Y., Hochberg, Y.: Controlling the false discovery rate: a practical and powerful approach to multiple testing. J. R. Stat. Soc. (1995).
20. van Rijthoven, M., Balkenhol, M., Siliņa, K., van der Laak, J., Ciompi, F.: HookNet: Multi-resolution convolutional neural networks for semantic segmentation in histopathology whole-slide images. Med. Image Anal. 68, 101890 (2021).

# Supplementary Materials & Methods

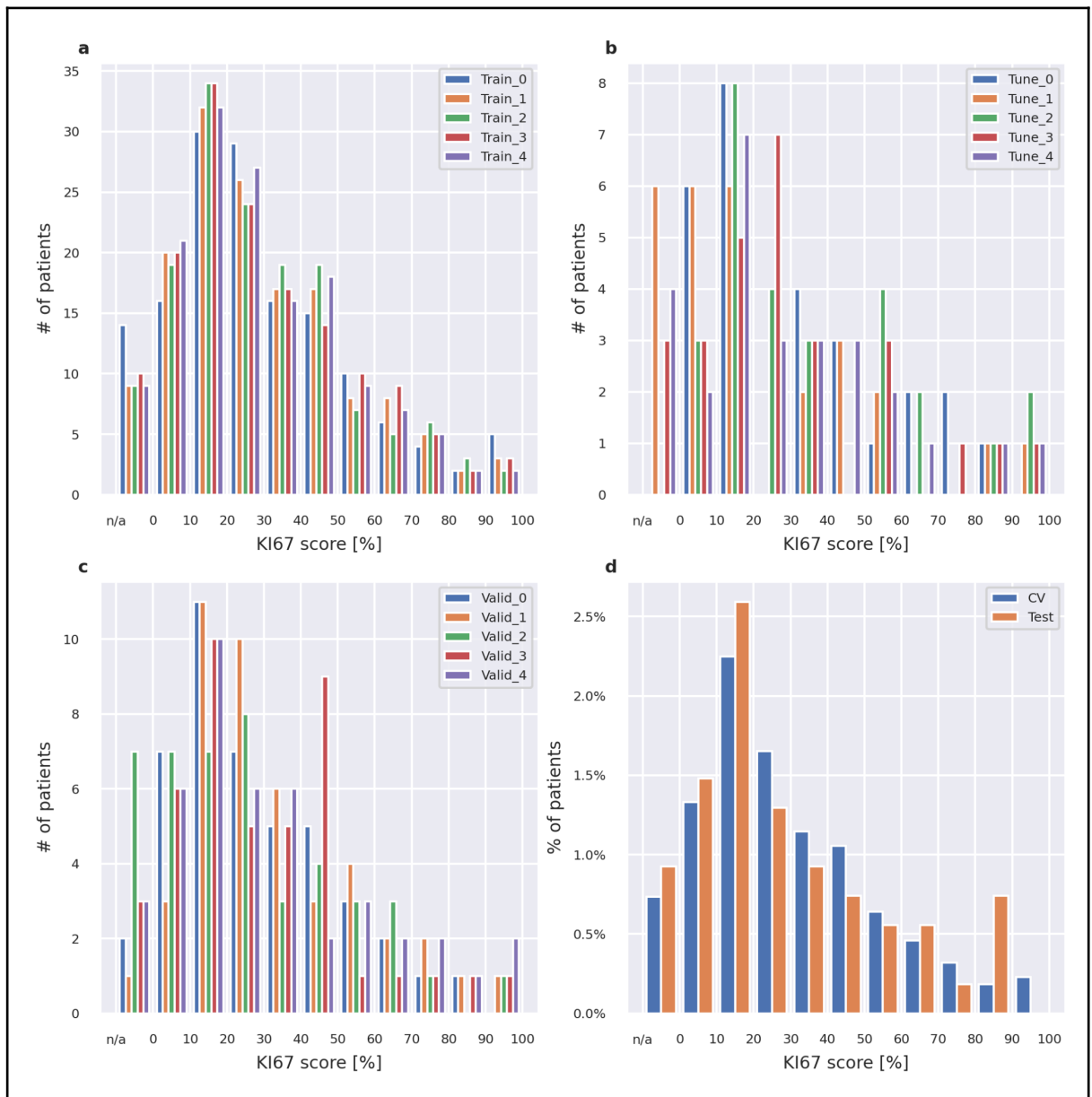

**Supplementary Figure 1.** *Distribution of Kl67-scores in different subsets of the data. a) shows the distribution of Kl67 scores in the training folds of the development data set, whereas b) and c) depict the distribution of Kl67-scores in the corresponding tune and validation folds respectively. d) shows a comparison of the CV/development data set and the test set.*

| label | lr | model | youden | auc | dice | accuracy | f1 |
|---|---|---|---|---|---|---|---|
| label_registered | 0.001 | resnet | 0.391 | 0.963 | 0.84 | 0.906 | 0.84 |
| label_registered | 0.001 | inception | 0.427 | 0.961 | 0.84 | 0.907 | 0.84 |
| label_registered | 0.0005 | inception | 0.424 | 0.961 | 0.84 | 0.907 | 0.84 |
| label_registered | 0.0005 | resnet | 0.386 | 0.96 | 0.835 | 0.903 | 0.835 |
| label_manual | 0.001 | inception | 0.457 | 0.976 | 0.887 | 0.928 | 0.887 |
| label_manual | 0.001 | resnet | 0.447 | 0.976 | 0.886 | 0.927 | 0.886 |
| label_manual | 0.0005 | resnet | 0.473 | 0.975 | 0.885 | 0.926 | 0.885 |
| label_manual | 0.0005 | inception | 0.443 | 0.975 | 0.881 | 0.924 | 0.881 |

**Supplementary Table 1.** *Hyperparameters and their corresponding performance metrics and thresholds.*

| | AUROC | Dice | Jaccard | Accuracy | F1 | Specificity | Sensitivity/ Recall | Precision |
|---|---|---|---|---|---|---|---|---|
| IHC-specific label - registered label | 0.954 [0.904, 0.983] | 0.948 [0.907, 0.974] | 0.948 [0.907, 0.974] | 0.882 [0.784, 0.947] | 0.948 [0.907, 0.974] | 0.879 [0.765, 0.95] | 0.918 [0.857, 0.963] | 0.964 [0.938, 0.981] |
| KI67-score - IHC-specific label | 0.175 [-0.115, 0.452] | 0.267 [-0.03, 0.538] | 0.269 [-0.026, 0.542] | -0.029 [-0.348, 0.285] | 0.268 [-0.027, 0.539] | -0.226 [-0.538, 0.109] | 0.389 [0.138, 0.605] | 0.149 [-0.161, 0.441] |
| KI67-score - registered label | 0.221 [-0.06, 0.482] | 0.286 [0.001, 0.547] | 0.286 [0.006, 0.543] | -0.066 [-0.369, 0.246] | 0.284 [-0.001, 0.544] | -0.257 [-0.547, 0.068] | 0.477 [0.23, 0.681] | 0.124 [-0.174, 0.417] |

**Supplementary Table 2.** *Spearman correlations between performance metrics for models trained with IHC-specific and registered annotations, as well as correlations of clinical KI67-scores with model performances for these two modelling approaches.*